# Convolutional neural network based deep-learning architecture for intraprostatic tumour contouring on PSMA PET images in patients with primary prostate cancer– a multicentre study including validation with histopathology standard of reference

Dejan Kostyszyn [a,b] *, Tobias Fechter [a,c,d] *, Nico Bartl [c,d], Anca L. Grosu [c,d], Christian Gratzke [e], August Sigle [e], Michael Mix [f], Juri Ruf [f], Thomas F. Fassbender [f], Selina Kiefer [g], Alisa S. Bettermann [d], Nils H. Nicolay [c,d], Simon Spohn [c,d], Maria U. Kramer [d], Peter Bronsert [g, h], Hongqian Guo, Xuefeng Qiu [h], Feng Wang [i], Christoph Henkenberens [j], Rudolf A. Werner [k], Dimos Baltas [a,d], Philipp T. Meyer [f], Thorsten Derlin [k], Mengxia Chen [h] **, Constantinos Zamboglou [c,d,l] **

* and ** authors contributed equally

(a) Division of Medical Physics, Department of Radiation Oncology, Medical Center – University of Freiburg, Faculty of Medicine. University of Freiburg, Germany
(b) Faculty of Engineering. University of Freiburg, Germany
(c) German Cancer Consortium (DKTK). Partner Site Freiburg, Germany
(d) Department of Radiation Oncology, Medical Center – University of Freiburg, Faculty of Medicine. University of Freiburg, Germany
(e) Department of Urology, Medical Center – University of Freiburg, Faculty of Medicine. University of Freiburg, Germany
(f) Department of Nuclear Medicine, Medical Center – University of Freiburg, Faculty of Medicine. University of Freiburg, Germany
(g) Institute for Surgical Pathology, Medical Center – University of Freiburg, Faculty of Medicine. University of Freiburg, Germany
(h) Department of Urology, Medical School of Nanjing University, Affiliated Drum Tower Hospital, China
(i) Department of Nuclear Medicine, Medical School of Nanjing University, Affiliated Drum Tower Hospital, China
(j) Department of Radiation Oncology, Hannover Medical School, Hannover, Germany
(k) Department of Nuclear Medicine, Hannover Medical School, Hannover, Germany
(l) Berta-Ottenstein-Programme, Faculty of Medicine, University of Freiburg, Germany

**Corresponding author**

Dr. Constantinos Zamboglou, M.D.

Department of Radiation Oncology, Medical Center – University of Freiburg, Robert-Koch Straße 3, 79106 Freiburg

Email: constantinos.zamboglou@uniklinik-freiburg.de

Telephone: 0049-761-270-94010, Fax: 0049-761-270-94620

**Abstract**

**Rationale:** Accurate delineation of the intraprostatic gross tumour volume (GTV) is a prerequisite for individualized diagnostic and therapeutic approaches in patients with primary prostate cancer (PCa). Several studies showed that prostate-specific membrane antigen positron emission tomography (PSMA-PET) may outperform magnetic resonance imaging in GTV detection. However, visual GTV delineation is underlying interobserver heterogeneity and is time consuming. The aim of this study was to train and to validate the performance of a convolutional neural network (CNN) for automated segmentation of intraprostatic tumour (GTV-CNN) in PSMA-PET images.

**Material and Methods:** The CNN (3D U-Net architecture) was trained on [$^{68}$Ga]PSMA-PET images of 152 patients from two different institutions (Freiburg and Nanjing) and the training labels were generated manually using a validated contouring technique. The CNN was tested on two independent internal (cohort 1: [$^{68}$Ga]PSMA-PET, n=18 and cohort 2: [$^{18}$F]PSMA-PET, n=19) and one external (Hannover, cohort 3: [$^{68}$Ga]PSMA-PET, n=20) test-datasets. Accordance between manual contours and GTV-CNN was assessed with Dice-Sørensen coefficient (DSC) and two distance based metrics. Additionally, sensitivity and specificity were calculated for the two internal test-datasets by using co-registered whole-mount histology as reference material.

**Results:** Median DSCs for cohorts 1-3 were 0.84 (range: 0.32-0.95), 0.81 (range: 0.28-0.93) and 0.83 (range: 0.32-0.93), respectively. The good DSC results were supported by distance based metrics. Sensitivities and specificities for GTV-CNN were comparable with manual expert contours: 0.98 and 0.76 (cohort 1) and 1 and 0.57 (cohort 2), respectively. Computation time was around 6 seconds for a standard dataset (288x288x426 voxels).

**Conclusion:** The application of a CNN for automated contouring of intraprostatic GTV in [$^{68}$Ga] and [$^{18}$F]PSMA PET images resulted in a high concordance with expert contours and in high sensitivities and specificities in comparison with histology reference. After further validation, this robust, accurate and fast technique may be implemented for individualized diagnostic and therapeutic concepts in patients with primary PCa. The trained model and the study's source code are available in an open source repository.

**Introduction**

Prostate cancer (PCa) is the most common tumour entity for men in North America and Europe in 2019 (1). In patients with newly diagnosed PCa, accurate contouring of the intraprostatic gross tumour volume (GTV) is mandatory for successful fusion-biopsy guidance (2). Additionally, focal therapy approaches such as focal dose escalation in radiotherapy, high intensity focused ultrasound or irreversible electroporation (3) rely on an accurate definition of the intraprostatic GTV.

Prostate-specific membrane antigen positron emission tomography (PSMA-PET) has recently been established for initial staging in primary PCa patients by providing an excellent diagnostic accuracy in detection of lymph node and bone metastases (1)). It is also increasingly used for intraprostatic staging in these patients in order to improve tumour lesion detection (4-7), focal therapy guidance (8) and non-invasive PCa characterization (9) compared to magnetic resonance imaging (MRI) as the current standard of care. Most of the studies evaluated [$^{68}$Ga]PSMA-11 as radiopharmaceutical. However, [$^{18}$F]PSMA-1007 is increasingly used for PMSA-PET imaging and Kuten et al. reported that [18F]PSMA-1007 may detect additional low-grade lesions in direct comparison with [$^{68}$Ga]PSMA-11 (10). In a recent work by our group manual and semi-automatic contouring approaches for [68Ga]-PSMA-PET images were validated based on co-registered histopathology reference material (11). Although good results (sensitivity and specificity > 80%) were obtained for most of the contouring approaches, some methodologies showed a rather poor performance (sensitivity and specificity < 70%). This is in line with a dice-index (DSC) varying between 0.56-0.8 for the manual contours, which indicates that PSMA-PET based GTV-definition underlies a substantial interobserver variability. Actually, no validated contouring technique for [18F]-PSMA-PET was proposed so far.

The implementation of an automatic segmentation algorithm may enhance intraprostatic GTV-delineation in PSMA-PET images by extending the two main limits of conventional contouring approaches: interobserver heterogeneity and expenditure of time. With the rise of deep learning in the recent years convolutional neural networks (CNNs) based algorithms achieved remarkable results handling this task. In a work by Zhao et al. the pelvic PCa tumour burden in [$^{68}$Ga]-PSMA-PET images was accurately detected by a CNN with 99% precision (12). Although several works already reported the excellent performance of CNNs in prostatic gland delineation on CT images (13) or PCa diagnosis based on multiparametric magnetic resonance images (14) the usage of CNNs for intraprostatic GTV contouring in PSMA-PET was not examined yet.

The aim of this work is to examine the capabilities of CNNs for intraprostatic GTV contouring in [$^{68}$Ga]- and [18F]-PSMA-PET images. Models were trained on two independent cohorts (Freiburg and Nanjing) with [$^{68}$Ga]-PSMA-PET images using a validated technique for manual intraprostatic GTV contouring (11). The network was thoroughly evaluated on external (Hannover, 1 cohort: [$^{68}$Ga]-PSMA-PET/CT scans) and internal (2 cohorts from Freiburg: [$^{68}$Ga] or [$^{18}$F]PSMA-PET/CT scans) cohorts as well as on co-registered whole-mount histopathology reference.

**Methods and Materials**

**Patients**

The current analysis evaluated data from 209 patients with primary PCa from three different centres (centre 1: Freiburg, centre 2: Nanjing, centre 3: Hannover). Inclusion criteria were histologically proven primary adenocarcinoma of the prostate and no local or systemic treatment prior to PSMA-PET imaging. The local ethics committees from all three centres approved the study and written consent was waived due to the retrospective character of the study. Please see table 1 for a detailed description of the patients` characteristics.

The training cohort consisted in total of 152 (107 centre1 and 45 centre 2) patients with 68Ga]PSMA-11 PET/CT scans. The CNN was tested on 3 independent test datasets: 18 patients from centre 1 ([68Ga]PSMA-11, cohort 1), 19 patients from centre 1 ([$^{18}$F]PSMA-1007, cohort 2) and 20 patients from centre 3 ([$^{68}$Ga]PSMA-11, cohort 3) Further, the CNN contours were assessed by coregistered histopathology information in both internal validation cohorts.

**PET/CT Imaging**

**PET/CT Imaging: Centre 1**

A detailed description of the radiolabelling protocol of [$^{68}$Ga]-PSMA-11 and [$^{18}$F]PSMA-1007 from centre 1 can be found in Zamboglou et al (15) and in Cardinale et al. (16), respectively. One hour ([$^{68}$Ga]-PSMA-11) and two hours ([$^{18}$F]PSMA-1007) after intravenous tracer injection, all patients underwent whole body PET scan. Protocols were acquired on three different cross-calibrated Philips scanners: GEMINI TF TOF 64, GEMINI TF 16 Big Bore and Vereos. All scanners fulfilled the requirements indicated in the European Association of Nuclear Medicine (EANM) imaging guidelines and obtained EANM Research Ltd. (EARL) accreditation during acquisition (17, 18). All scanners resulted in a PET image with a voxel size of 2 x 2 x 2 mm$^3$. Images were normalized to decay corrected injected activity per kg body weight (standardized uptake values, SUV in [g/ml]). As a result of the EARL accreditation process, comparable SUV parameters were derived.

**PET/CT Imaging: Centre 2**

[$^{68}$Ga]PSMA-11 was synthesized using an ITG semiautomated module as described previously (19). One hour before scanning, all patients were intravenously injected with [$^{68}$Ga]PSMA-11. PET/CT was performed on an uMI 780 PET/CT scanner (United Imaging Healthcare). A CT scan (130 keV, 80 mAs) and a static emission scan, corrected for dead time, scatter, and decay, were acquired from the vertex to the proximal legs. The scanner resulted in a PET image with a voxel size of 2.3 x 2.3 x 2.7 mm. A resampling step was performed to obtain a PET image voxel size of 2 x 2 x 2 mm (tri-linear interpolation in plastimatch v1.8.0) before training of the CNN. Expert contours of intraprostatic GTV and prostate contours were resampled with nearest neighbor interpolation (plastimatch v1.8.0).

**PET/CT Imaging: Centre 3**

[68Ga]-PSMA-11 was synthesized as described previously (20). All studies were acquired using a Biograph mCT 128 Flow scanner (Siemens, Knoxville, USA) one hour after tracer injection, as described previously (20).
PET images had a voxel size of 4.1 x 4.1 x 5 mm. Validation was performed with the original data and with three different resampling methods to obtain a PET image voxel size of 2 x 2 x 2 mm.

**Histopathology and PET/CT image coregistration**

For 29 patients from centre 1 (cohort 1: n=18 and cohort 2: n=11) the three-dimensional (3D) distribution of the intraprostatic GTV was obtained by histology information from prostatectomy specimen using an in-house coregistration protocol, as previously described (13, 21). After formalin fixation, the resected specimen underwent an ex-vivo CT scan using a customized localizer and whole-mount step sections were cut every 4 mm using an in-house cutting device. Staining with hematoxylin and eosin was performed via routine protocols and PCa tissue in histology was delineated by pathologists. Histology slices were registered on ex-vivo CT images and PCa contours were transferred onto the CT images. The contours were interpolated to create a model of the 3D distribution of PCa in histology (GTV-Histo). Ex-vivo CT (including GTV-Histo contours) was registered on in-vivo CT (from PSMA-PET/CT scans) by manual coregistration including elastic deformations. The final alignment of in-vivo CT and PET scans was based on the hardware coregistration of the hybrid PET/CT scanners.

**Contouring of PSMA PET/CT images**

For all patients, GTVs based on [$^{68}$Ga]PSMA-PET images were manually delineated by two experienced readers (GTV-Exp) in consensus using 3D Slicer (Version 4.10.0) as proposed previously by our group (11). Any monofocal or multifocal uptake greater than adjacent background uptake in more than one slice within the CT-defined prostate gland was defined as presence of PCa. GTVs were delineated manually in every single slice using inverted grey color scale for display, thresholded with a SUVmin and max of 0 and 5, respectively. GTV-PET on [$^{18}$F]PSMA-PET images was obtained by using the same procedure. Additionally, threshold based contouring with 30% of intraprostatic SUVmax was applied (GTV-30%) (4) for the patients with histopathology reference in test cohorts 1 and 2.

The contour of the prostatic gland on corresponding CT images was delineated by an experienced radiation oncologist for all patients.

**Preprocessing**

The data were converted from the digital imaging and communications in medicine (DICOM) standard to the nearly raw raster data (NRRD) format, cropped to a size of 64x64x64 voxels in all dimensions and normalized with $x_i' = \frac{x_i - \overline{x}}{\sigma}$, where $x_i$ is the PET data of patient $i$, $\overline{x}$ the arithmetic mean and $\sigma$ the standard deviation within all cropped datasets in the corresponding training cohort.

As it is not possible to accurately differentiate between prostatic tissue and surrounding tissues solely based on PET and since differences in the contours due to interobserver variability could mislead the network, only delineations inside the prostatic gland contour were used for computations.

To investigate the impact of a voxel size different from the training voxel size and the usage of different interpolation algorithms we used the PETs from centre 3 in four different ways. First, the original data was fed to our network. In a second setting, the PETs from centre 3 were resampled to a resolution of 2 x 2 x 2 mm with three different methods (SimpleITK v1.2.4): B-spline interpolation order 3, tri-linear interpolation and Gaussian interpolation. Prostate contours and ground truth GTVs were resampled with nearest neighbor interpolation.

**Convolutional neural network**

According to the majority of successful end-to-end segmentation methods for volumetric medical image segmentation tasks (17, 22), the current work was based on a 3D variant of the U-Net architecture. The chosen network consists of 3 down sampling steps with max-pooling, 3 upsampling steps with transposed convolution layers (kernel size: 2x2x2, stride: 2, padding: 1) and skip connections by concatenation. The 18 convolution blocks consist of 3 x 3 x 3 convolutions with stride and padding of 1, followed by Batch Normalization (BN) and ReLU activation, except for the last convolution where 1 x 1 x 1 convolution without padding, BN and Sigmoid activation function were used. An argmax function over the final feature map formed the predicted GTV. An illustration of the network can be found in Figure 1. The network weights were optimized using ADAM (23).

**Convolutional neural network: Training**

The 152 patients of the training cohort were further split into training and evaluation cohort of 142 and 10 patients, respectively. The evaluation cohort was used for optimizing the CNN's hyper-parameters during the training process. As input the CNN received a concatenation of the patients' PET and prostate contour. Hyper-parameter optimization was done using a grid search. The best performing setting was achieved with Adam $\beta_1 = 0.9 \wedge \beta_2 = 0.999$, a learning rate of 0.0001 and training for 1019 epochs (an epoch means iterating over all training samples once) with a dice loss: $diceloss(X,Y) = 1 - \frac{2\sum_{l=1}^{|L|} w_l \sum_n y_{ln} x_{ln}}{\sum_{l=1}^{|L|} w_l \sum_n y_{ln} + x_{ln}}$ for $|L|$ number of labels, N image elements $x_{0,\dots,N} \in X, y_{0,\dots,N} \in Y$ and without weighting the label classes $w_l = 1$. Data augmentation was performed during grid search but achieved worse results than the above mentioned setting without augmentation. Consequently no data augmentation was used for further analyses.

In Figure 2 a visualization of the training and evaluation curves can be seen. As both loss curves decrease almost simultaneously and the distance between them stays constant we assume that the network generalizes well and overfitting is unlikely.

**Convolutional neural network: Evaluation**

Due to the mentioned reasons in 3.5 only the output of the CNN inside the contour of the prostatic gland was considered.

We assessed the agreement between GTV-Exp and GTV-CNN at voxel level using the Dice-Sørensen coefficient (DSC). When applied at voxel level, this index is identical to the kappa index (8). As volume-based metrics (like DSC) show a lower sensitivity to errors where outlines deviate and the volume of the erroneous region is small compared to the total volume, we considered also distance-based metrics like the Hausdorff distance (HD) and the average symmetric surface distance (ASSD). Additionally, we calculated the sensitivity and specificity for all GTVs based on the histology standard of reference data as performed previously by our group (13). The prostate in each CT slice (from PSMA-PET/CT scans) was divided into four equal segments and the analysis was performed visually using the GTVs obtained. A median of 52 segments (range: 20-64) per patient were analysed.

In a last experiment we were interested whether there are clinical factors that might impact the CNN performance by influencing the SUV distribution (PSA values and Gleason score) or by neighborhood to bladder signal (localization). Patients from cohorts 1 and 3 (resampling: tri-linear interpolation) were pooled and a logistic regression was performed to assess the impact of initial PSA values, Gleason score, cT stage and tumour localization on the concordance of GTV-CNN and GTV-Exp in terms of DSC. The median DSC of pooled cohorts was used as cut-off point.

**Convolutional neural network: Implementation**

The network was implemented with pytorch 1.3.1 and torchvision 0.4.2. Gradients for backpropagation were calculated with the pytorch autograd library which keeps track of all operations and builds a computational tree.

For further details please see the provided code on GitLab: (https://gitlab.com/dejankostyszyn/prostate-gtv-segmentation)

Computations were done on a PC with a Intel ® Xeon ® Silver 4114 CPU @ 2.20GHz, 64 GB memory and a NVIDIA ® Quadro ® P5000, 16GB, 4DP, DL-DVI-D GPU.

**Statistical analysis**

The statistical analysis was performed with MedPy's package 'Metric Measures' v0.4.0 and GraphPad Prism v8.1.0 (GraphPad Software). Pairwise comparisons (DSC, ASSD, HD, sensitivity and specificity) were performed with the Wilcoxon matched-pairs signed rank or Friedman test. Non-pairwise testing was performed with Mann-Whitney test (initial PSA) and Chi-square test (cT stage and Gleason score). The tests were chosen due to non-normal distribution (Shapiro-Wilk test) of the data. Furthermore, a binary logistic regression analysis with pooled cohorts 1 and 3 was performed to assess the impact of basic clinical parameter on DSC between GTV-Exp and GTV-CNN. The confidence alpha was set to 5% for all analyses.

**Results**

**Test results [$^{68}$Ga]-PSMA-11 PET**

GTV-CNN was compared to GTV-Exp on internal and external testing cohorts. On the internal datasets (cohort 1) the network yielded median DSC, median HD and median ASSD of 0.84 (range: 0.32-0.94), 4mm (range: 1.41-10) and 0.61mm (range: 0.24-1.46), respectively (Table 2). CNN performance was further assessed by correlation with histology reference (Figure 3) and median sensitivity and specificity of 0.98 (range: 0.38-1) and 0.76 (range: 0.13-1) were observed.

The achieved sensitivity and specificity was comparable to GTV-Exp and GTV-30% (Figure 4). The median absolute volumes of the GTVs were: 10.7ml (range: 0.7-101) for GTV-CNN, 11.8ml (range: 0.8-75) for GTV-Exp, 8ml (range: 2.2-41) for GTV-30% and 10.4ml (range:

1.6-103) for PCa volume in histology reference. No significant differences between absolute volume of GTV-CNN and the three other volumes were observed. The GTV-CNN encompassed in median 26.6% of the prostatic gland.

Patients in the external test cohort (cohort 3) had statistically significant differences between Gleason scores but not between initial PSA values and cT stage (Table 1). Comparison between GTV-CNN and GTV-Exp was performed firstly on non-resampled and secondly on resampled PET images (Table 2). Friedman test revealed statistically significant ($p<0.01$) differences in DSC, HD and ASSD among the different pre-processing procedures and no pre-processing. Post-hoc analyses revealed no statistically significant differences between the three interpolation approaches ($p>0.05$). Best results were achieved by using tri-linear interpolation with median DSC 0.83 (range: 0.32-0.93), median HD 4.12mm (range: 2.01-22.36) and median ASSD 0.46mm (range: 0.28-1.61), respectively.

In logistic regression analysis with pooled cohorts 1 and 3 no clinical parameter had a statistically significant impact on DSC between GTV-Exp and GTV-CNN (supplementary Table 1).

### Test results [$^{18}$F]-PSMA-1007 PET

The trained CNN was also tested on 19 patients from cohort 2. Considering GTV-PET for comparison median DSC, median HD and median ASSD were 0.81 (range: 0.28-0.93), 5mm (range: 1.41-8.49) and 0.51mm (range: 0.26-1.57), respectively (Table 2). Considering histology as standard of reference calculated sensitivity and specificity were 1 (range: 0.86-1) and 0.57 (range: 0.12-1). GTV-CNN had a significant higher sensitivity than GTV-30% ($p=0.01$) but not than GTV-Exp ($p=0.48$). No statistically significant differences in specificity ($p>0.05$) were observed between GTV-CNN and the other two GTVs. In the 11 patients with co-registered histology information median tumour volume was 3.5ml (range: 0.3-24.4) for histology reference, 8.5ml (range: 1.9-38) for GTV-CNN, 3 ml (range: 0.6-21.5) for GTV-30% and 7.2ml (range: 1.2-36) for GTV-Exp. GTV-CNN was statistically significant larger ($p>0.05$) than all three other volumes ($p<0.05$) and encompassed in median 32% (range: 9-86) of the prostatic gland.

### Computation time

For internal validation cohorts ([$^{18}$F] and [$^{68}$Ga]PSMA) the segmentation of the intraprostatic tumour of one patient took in median 6 and 6.28 seconds, respectively, for the complete process, including loading the PET and prostate contour from hard drive and storing the predicted GTV onto hard drive. This process took 23.3-27.8 seconds (in dependency of the used interpolation technique) for external validation cohort, because the data was scaled to 443% of the internal cohort's data volume after resampling. A single forward pass through the CNN took less than a second (approx. 3 µs) for all cohorts. For further information please see Table 2.

### Discussion

Implementation of automatic GTV-segmentation approaches based on CNN algorithms have already been introduced for several other tumour sites like gliomas (18, 24) or lung cancer (21). Although several studies achieved promising results by using CNNs for auto segmentation of the prostatic gland in mpMRI or CT there is limited evidence on the segmentation of the intraprostatic GTV(25). To the best of our knowledge our study is the first study analyzing CNNs for intraprostatic GTV delineation based on PET images. We chose PSMA-PET images since several studies reported that PSMA-PET outperforms the current standard of care (multiparametric magnetic resonance imaging) for intraprostatic tumour detection (4-7). Consequently, the use of PSMA-PET imaging for initial staging (26) and intraprostatic GTV detection and contouring (27, 28) has been established in the last years and several studies suggested its implementation for treatment individualization in primary PCa patients (29, 30). However, all previous studies used manually or semi-automatically created contours for intraprostatic GTV contouring which may be impeded by low sensitivity/specificity and interobserver heterogeneity (11). It should be also considered, that manual contouring of GTV in the prostate is time consuming and may bind human resources. Obviously, a fast, robust and accurate workflow for intraprostatic GTV contouring is a prerequisite for a broader deployment of PSMA PET-based procedures targeting the prostatic gland. In this work we were able to show for the first time that CNN have the ability to delineate the intraprostatic GTV on PSMA-PET with accuracy comparable to human experts within seconds. Thus, it is likely that PSMA-PET/CT imaging in combination with CNN-based intra- and extraprostatic (12) tumour detection and segmentation may provide a "one-stop shop" tool for tailoring individualized treatment approaches for primary PCa patients.

In our study the CNN was trained with [$^{68}$Ga]-PSMA-PET data of 152 patients from two different institutions using a validated approach for GTV delineation. The CNN performance for [$^{68}$Ga]PSMA-PET images was tested on two independent datasets as well as using histology reference material. In comparison with manually delineated expert contours very good DSC values (>0.8) were observed for both validation datasets. Bravaccini et al. reported that the PSMA expression correlates with the Gleason score (31) and the two test cohorts had statistically significant differences in Gleason score in biopsy probes. The good agreement between GTV-CNN and GTV-Exp in both test cohorts suggests that the CNN performance is independent of the Gleason score. In rare cases a high HD was observed despite a high DSC. This was the case when the main parts of CNN and expert GTVs overlapped, but small regions with a high distance to the main tumour, were diagnosed as malignant by the CNN, but not by the human experts. For example, in two patients of the internal validation cohort ([$^{68}$Ga]PSMA-PET) the CNN detected small (<0.5 mm in histology reference material) lesions which were missed by visual PET interpretation. This explains the slightly higher sensitivity of the CNN cohort 1 although the absolute GTV volumes were comparable between experts and CNN. Since HD is sensitive to outliers, we used ASSD as additional metric. In our tests all segmentations with regions to reassess had an ASSD > 1, while results < 1 could be considered as good segmentations. Hence, we recommend considering not just DSC, but also HD and ASSD to evaluate CNN contouring accuracies. In comparison with histology reference GTV-CNN achieved excellent sensitivity and good specificity in [$^{68}$Ga]PSMA-PET images, which was comparable to manually delineated expert contours and threshold based contours using 30% of intraprostatic SUVmax. Additionally, the absolute volume of GTV-CNN was very similar to the histology reference volume suggesting a good coverage of the intraprostatic tumour mass by the CNN. Since GTV-CNN encompassed in median 26.6% of the prostatic gland, it is very likely that focal therapy approaches guided by CNN are feasible in most of the patients.

[$^{68}$Ga]PSMA-PET images of the external validation cohort were tested with and without previous resampling. Statistically significant differences were observed with better results for the resampled datasets. Hence, when using different datasets from different institutions a resampling of the images to the same voxel size of the training data set should be performed in order to enhance the CNN performance. In direct comparison between three different interpolations for resampling no statistically significant differences were observed. However, considering a decent better performance for tri-linear interpolation this method should be considered for future studies. Consequently, the automatic GTV delineation in [$^{68}$Ga]PSMA-PET images with CNNs is a promising tool for intraprostatic tumour segmentation offering a robust, accurate and fast alternative to visual PET image interpretation. However, in some patients discrepant results between CNN and manual contours were observed. Moreover, PET signal from the adjacent bladder may mislead the CNN in contouring of PCa lesions in the prostatic base. However, no clinical parameters like Gleason Score or tumor localization had an impact on the concordance between GTV-Exp and GTV-CNN in regression analysis. Thus, we strongly advise to perform a visual control of the CNN segmentations in every patient.

Interestingly, the CNN provided also a high concordance with expert contours (DSC>0.8) in contouring of [$^{18}$F]PSMA-PET images. This result should be interpreted with caution since no validated approach for contouring was applied. However, considering histology as standard of reference an excellent sensitivity was observed which was comparable to manual contours and better than threshold based contours. Taking into account the differences in physical properties and in bio distribution between both tracers this result is surprising. However, the specificity of GTV-CNN was low (<0.65) which is mainly explained by a significant overestimation of the tumour volume. Thus, the CNN may also be used for GTV contouring in [$^{18}$F]PSMA-PET images, especially in situations when a good coverage of the intraprostatic GTV is demanded and a high coverage of non-tumour bearing prostatic tissue is negligible. Surely, further studies implementing [$^{18}$F]PSMA-PET images and validated expert contours for training and testing are necessary to confirm this observation.

A limitation of our study is the relatively low number of patients used for testing the CNN (n=57) which is explainable by the elaborate co-registration protocol used in our study for two independent cohorts. We assume that the observed results are robust since we used different, independent datasets for evaluation and received comparable results for all of them. Another point that supports the robustness is that we did not notice any overfitting in the training process. Considering the high value of the two independent datasets used for testing the CNN, no additional approaches for validation were performed (e.g. k-fold cross-validation). A different potential limitation of our study is the uncertainty in correlation of PSMA PET images and histopathology slices (e.g. non-linear shrinkage of the prostate after prostatectomy). Thus, it could not be excluded that moderate or low coverage of PCa in histology by the PET-derived GTVs is a consequence of mismatch in coregistration or incomplete histopathological coverage instead of poor tracer or CNN performance. Additionally, Gibson et al. proved that pathologists tend to underestimate the true extent of disease (32). However, as the calculation of sensitivities and specificities was not performed on a voxel-level but on a less stringent slice by slice level, we consider the potential resulting bias negligible. Another issue is that, mainly PCa patients with intermediate- or high-risk PCa are referred for PSMA-PET imaging. As a result, the findings from our study are only representative for intermediate- and high-risk PCa patients. Consequently, further studies need to proof our results for low-risk PCa patients. Very recently a new CNN architecture (modified U-NET) was introduced (33) in order to overcome potential limitations in segmentation of small lesions. Future studies should assess whether implementation of this new approach further increases the results.

In conclusion, our study presents a CNN for automated contouring of intraprostatic GTV in [$^{68}$Ga] and [$^{18}$F]PSMA-PET images resulting in a high concordance with expert contours and in high sensitivities and specificities in comparison with histology reference. Likewise, CNN-based GTV delineation is a promising and fast alternative to visual PET image interpretation offering a high accuracy. The link to the code and trained model of the CNN can be found in chapter 3.5.4. and may be used for individualized treatment (e.g. focal therapy) and diagnostic (e.g. radiomic feature extraction or fusion-biopsy) concepts in patients with primary PCa on two ways: (1) before visual image interpretation the CNN may provide a proposal and (2) after visual image interpretation the CNN may be used for quality assurance. However, we strongly emphasize that a visual control of the CNN contours by experienced experts is still obligatory in both approaches. Additionally, we will further train and validate the network and will offer updated versions of the underlying code.

## Abbreviations

ASSD: average symmetric surface distance; CNN: convolutional neuronal network; DSC: Dice-Sørensen coefficient; GTV:gross tumor volume; HD: Hausdorff distance; mpMRI: multiparametric magnetic resonance imaging; PCa: Prostate cancer; PSMA-PET: prostate specific membrane antigen positron emission tomography; SUV: standardized uptake values

## Disclosures

All authors confirm that no conflicts of interest (financial or otherwise) exist, that may directly or indirectly influence the content of the manuscript submitted.

## Tables

**Table 1. Patient characteristics**

|  | **Centre 1** | **Centre 2** | **Centre 3** |
|---|---|---|---|
| Median Age in years (range) | 70 (48-88) | 69 (55-84) | 71.5 (59-84) |
| Median PSA in ng/ml (range) | 13.1 (4.4-218) | 13.3 (4.04-110) | 12.8 (1.91-108.10) |
| Gleason Score, n and % | | | |
| 6 | 5 (3.4) | 4 (8.9) | 0 |
| 7a | 44 (30.1) | 12 (26.7) | 3 (15) |
| 7b | 43 (29.9) | 12 (26.7) | 3 (15) |
| 8 | 24 (16.7) | 10 (22.2) | 6 (30) |
| 9 | 19 (13.2) | 7 (15.5) | 8 (40) |
| unknown | 9 (6.3) | 0 | 0 |
| cT stage, n and % | | | |
| 2 | 89 (61.8) | 14 (30.8) | 6 (30) |
| 3 | 55 (38.2) | 31 (69.2) | 14 (70) |
| n patients with [$^{68}$Ga]PSMA-PET/CT, total | 125 | | 20 |
| n patients training cohort | 107 | 45 | |
| n patients validation cohort | 18 | | |
| n patients with histology reference | 18 | | |
| n patients with [$^{18}$F]PSMA-PET/CT, total | 19 | | |
| n patients validation cohort | 19 | | |
| n patients with histology reference | 11 | | |

Differences in the two test cohorts ([68Ga]-PSMA-PET images) from centre 1 and 3 were analyzed. No differences in initial PSA values and cT stage were observed (p>0.05). However, patients from centre 3 had statistically significant (p=0.035) higher Gleason scores than patients from centre 1.

**Table 2.**

| | DSC | | | HD (mm) | | | ASSD (mm) | | | Computation time (sec) | | |
|---|---|---|---|---|---|---|---|---|---|---|---|---|
| | Median | Min | Max | Median | Min | Max | Median | Min | Max | Median | Min | Max |
| **Internal validation [68Ga]PSMA-11** | 0.84 | 0.32 | 0.95 | 4.03 | 1.42 | 10.0 | 0.61 | 0.28 | 1.97 | 6.28 | 5.47 | 7.66 |
| **Internal validation [18F]PSMA-1007** | 0.81 | 0.28 | 0.93 | 5.0 | 1.41 | 8.49 | 0.5 | 0.26 | 1.82 | 6.00 | 3.53 | 9.2 |
| **External validation [68Ga]PSMA-11** | | | | | | | | | | | | |
| No resampling | 0.78 | 0.11 | 0.89 | 12.57 | 1.43 | 32.9 | 0.62 | 0.27 | 4.03 | 1.93 | 0.27 | 2.02 |
| B-spline interpolation | 0.82 | 0.39 | 0.92 | 5.83 | 2.45 | 22.36 | 0.55 | 0.32 | 2.1 | 27.79 | 10.54 | 30.91 |
| Tri-linear interpolation | 0.83 | 0.32 | 0.93 | 4.12 | 2.01 | 22.36 | 0.46 | 0.28 | 1.61 | 23.32 | 10.55 | 26.37 |
| Gaussian interpolation | 0.81 | 0.04 | 0.94 | 7.35 | 2.24 | 20.05 | 0.55 | 0.19 | 3.72 | 25.13 | 10.49 | 28.3 |

**Figures**

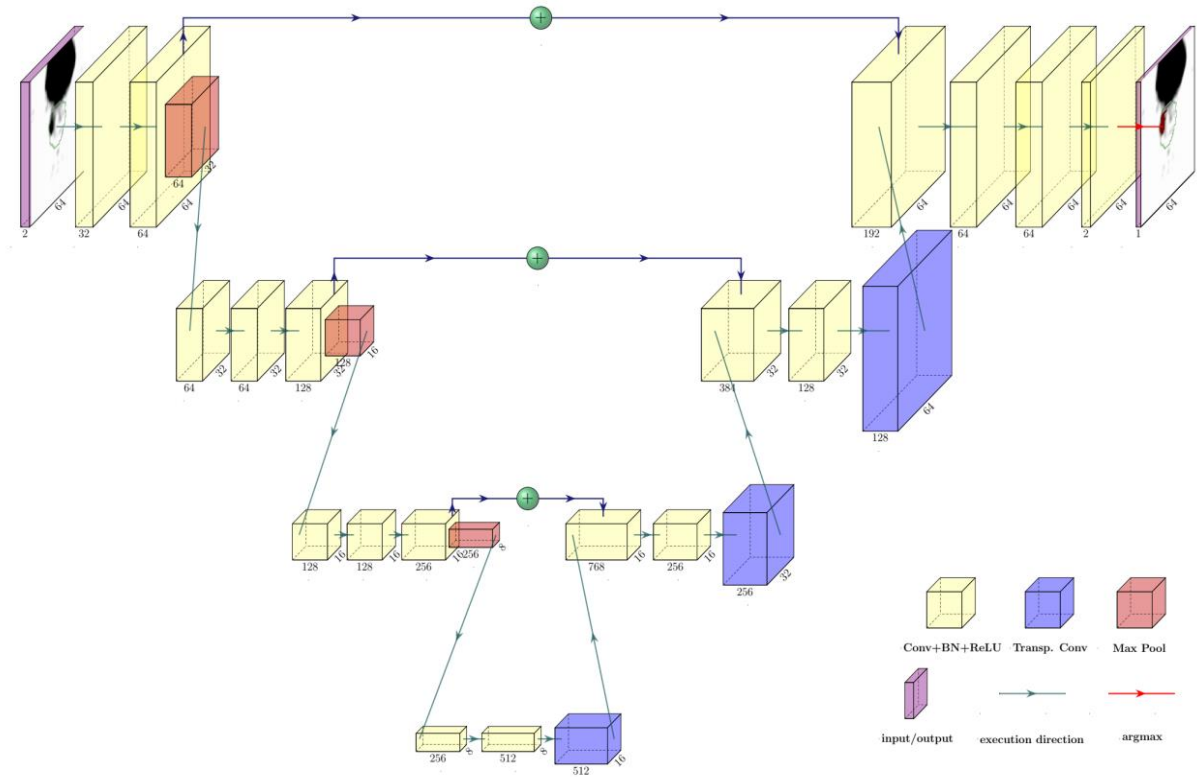

**Figure 1**. Illustration of the network. Numbers below layers represent the number of channels, numbers to the right of a layer represent the shape of the data cube after executing the layer's operations.

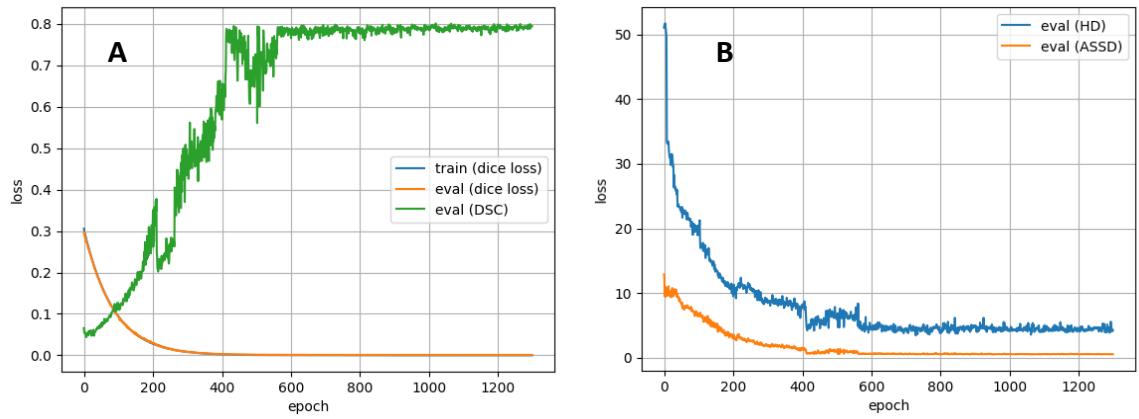

**Figure 2.** Visualization of the training and evaluation curves. Training and evaluation results as dice loss and Dice-Sørensen coefficient (DSC) in A as well as Hausdorff distance (HD) and average symmetric surface distance (ASSD) in B.

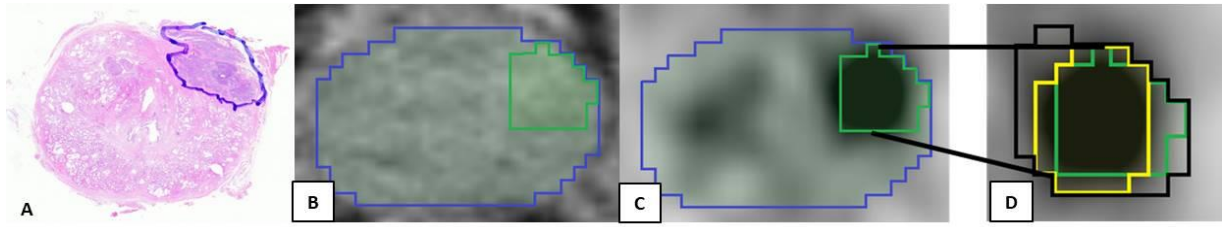

**Figure 3.** Histology reference and [$^{68}$Ga]-PSMA-11 PET/CT scan from a representative patient of cohort 1 are shown. A: Hematoxilin and eosin whole-mount prostate slide with marked PCa lesion in blue. B: Axial slide of CT-scan. Each CT slide was divided into 4 segments to calculate sensitivity and specificityfor each patient. C: Axial slide PET-scan (image scaling: SUVmin-max=0-5). D: Zoom of the PSMA-PET-positive Lesion. Blue contour: Prostate. Green contour: histology reference. Yellow contour: GTV-Exp. Black contour: GTV-CNN.

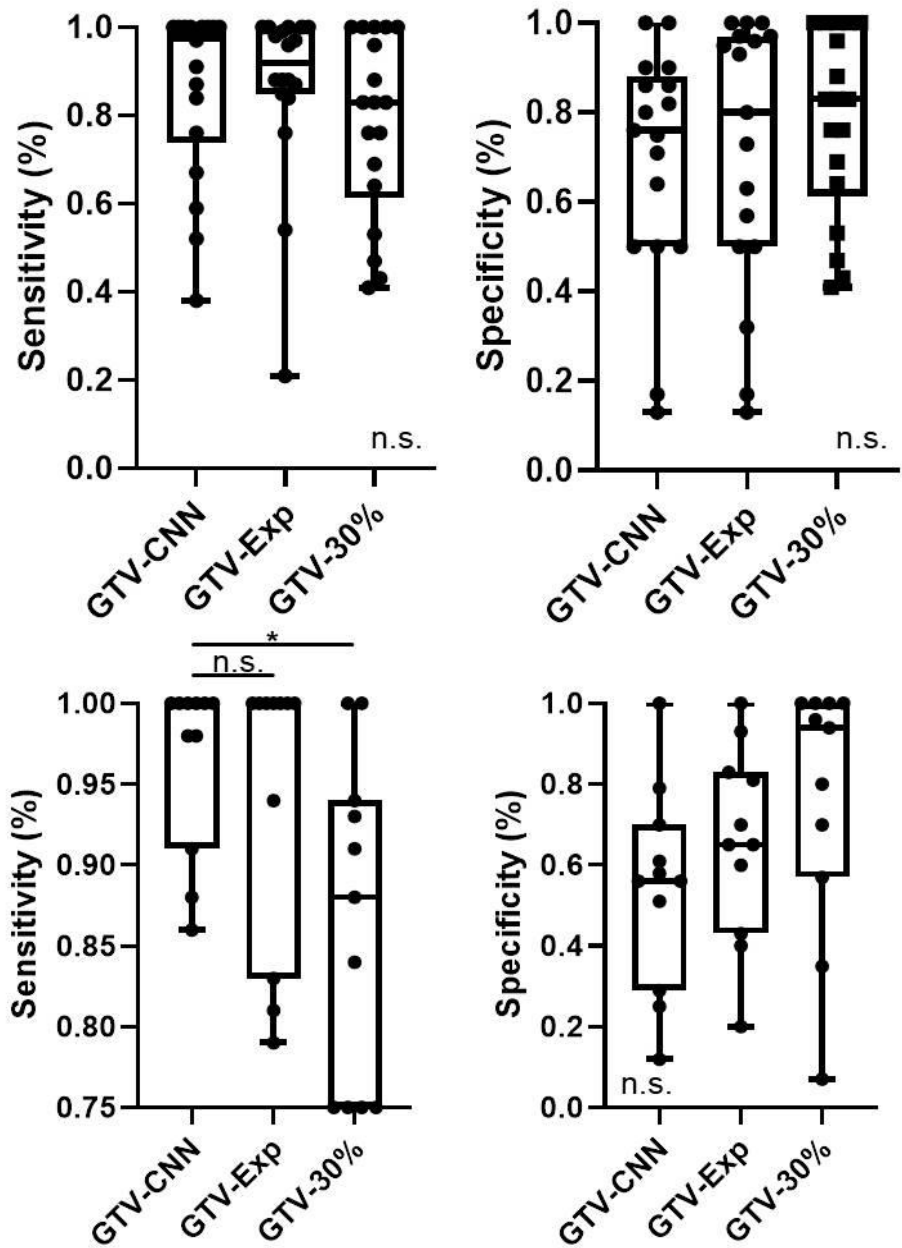

**Figure 4.** Comparison of specificity and sensitivity between GTV-CNN, GTV-Exp and GTV-30% based on histology reference in cohort 1 and 2. Box plots including median values, interquartile range and sensitivity/specificity for each individual patient are presented. Pairwise comparison was performed with Wilcoxon signed-rank test. Abbrevations: n.s.: not significant, *: p=0.05-0.01, CNN: convolutional neural network.

## Funding

This study was funded from the ERA PerMed call 2018 (BMBF)